# Extracting and Learning Fine-Grained Labels from Chest Radiographs

Tanveer Syeda-Mahmood, Ph.D, K.C.L Wong, Ph.D, Joy T. Wu, M.D., M.P.H, Ashutosh Jadhav, Ph.D, Orest Boyko, M.D. Ph.D
IBM Almaden Research Center, San Jose, California, USA

## Abstract

*Chest radiographs are the most common diagnostic exam in emergency rooms and intensive care units today. Recently, a number of researchers have begun working on large chest X-ray datasets to develop deep learning models for recognition of a handful of coarse finding classes such as opacities, masses and nodules. In this paper, we focus on extracting and learning fine-grained labels for chest X-ray images. Specifically we develop a new method of extracting fine-grained labels from radiology reports by combining vocabulary-driven concept extraction with phrasal grouping in dependency parse trees for association of modifiers with findings. A total of 457 fine-grained labels depicting the largest spectrum of findings to date were selected and sufficiently large datasets acquired to train a new deep learning model designed for fine-grained classification. We show results that indicate a highly accurate label extraction process and a reliable learning of fine-grained labels. The resulting network, to our knowledge, is the first to recognize fine-grained descriptions of findings in images covering over nine modifiers including laterality, location, severity, size and appearance.*

## 1. Introduction

Chest X-rays are the most common diagnostic exam in emergency rooms and intensive care units today. Due to the availability of large open source datasets[3,6] and labels for a small set of coarse-grained findings, such as opacities, they have become the subject of vigorous recent activity by researchers[1-6]. Deep learning algorithms have become the de facto approach to recognize the findings in these images. However, before such capabilities can be incorporated into clinical practices to produce automated preliminary reads, the models should be able to recognize not only a comprehensive and broad spectrum of radiographic findings but also describe them in a fine-grained fashion covering laterality, anatomical location, severity, appearance, etc. For example, saying "cardiomegaly" as the label for both the images in Figure 1b and 1c is not sufficient to describe these images as one constitutes a severe case (Figure 1c) and may need a more prompt attention. A full-fledged preliminary read radiology report describes various types of findings along with their positioning, laterality, severity, appearance characteristics, etc, as can be seen from a sample report shown in Figure 1a.

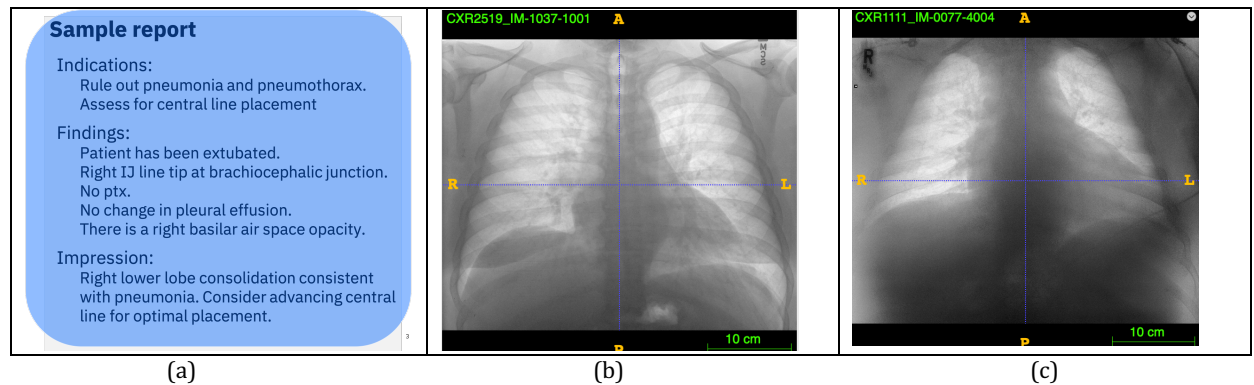

(a)  (b)  (c)

*Figure 1. Illustration of the need for capturing fine-grained descriptions of findings in deep learning models. (a) A sample radiology report. (b) an image depicting mild cardiomegaly, and (c) an image depicting severe cardiomegaly. Deep learning models should learn these detailed descriptions to enable automated preliminary reads by machines in future.*

Thus to capture realistic read scenarios, deep learning models should be trained on fine-grained labels. A number of recent approaches have attempted to take advantage of the associated radiology reports to automatically label the images[2,6,8]. However, they have been limited to a small number of core findings and have come under scrutiny due to inadequate verification. Complete labeling of images for all possible findings seen in that modality imaging is a challenging problem requiring the development of both vocabularies covering the findings and development of high precision and recall methods for extracting such labels from their associated radiology reports. Once the labels, novel deep learning architectures must still be developed that are suitable for such fine-grained label classification.

In this paper we present a comprehensive approach to extract and learn fine-grained labels from chest radiographs. Specifically, we develop a new descriptor for fine-grained labels utilizing the valid combinations of findings and their modifiers commonly found in radiology reports. We then develop a vocabulary-driven concept algorithm for automatically finding these core concepts and modifiers from sentences in radiology reports. The vocabulary for core findings and modifiers is derived from a clinician-curated chest X-ray lexicon. A phrasal grouping algorithm then associates detailed characterization modifiers with the relevant core findings in sentences . By applying the finding extractor algorithm on a large collection of radiology reports, we retain 457 fine-grained labels that have adequate image dataset to train deep learning models. We then develop a deep learning model that is designed for such fine-grained classification by exploiting an architecture based on feature pyramids and dilated convolution blocks. We show results that indicate a highly accurate label extraction process and a reliable learning of fine-grained labels. To our knowledge, the resulting network is the only network of its kind that can recognize fine-grained descriptions of findings in images covering several modifiers including laterality, location, severity, size and appearance.

## 2. Materials and Methods

### 2.1 Describing fine-grained labels

To describe findings in a fine-grained way, we adopt the following descriptor $F_i = <T_i|N_i|C_i|M_i^*>$ where $F_i$ is the fine-grained label, $T_i$ is the finding type, $N_i$ = yes|no indicates a positive or ruled out finding, $C_i$ is the core finding itself, and $M_i$ are one or more of the possible finding modifiers. In the pattern, each modifier is at its designated position separated by |. The finding types in chest X-rays are adequately covered by six major categories namely, anatomical findings, tubes and lines and their placements, external devices, viewpoint-related issues, and implied diseases associated with findings. By analyzing a large set of radiology reports, the set of relevant modifiers for each finding type were found to be slightly different as shown in Table 1.

The process used to derive the list of valid values for core finding labels and modifiers for each finding type was semi-automatic and involved a clinician-directed curation process. Specifically, a team of 4 clinicians (3 radiologists and 1 internal medicine doctor) used a combination of top-down and bottom-up processes to uncover the list of findings seen in AP chest radiographs and recorded them in a chest X-ray lexicon. The details of this multi-year lexicon construction effort is being described in a companion submission, so we will briefly outline it here to add relevant context to our fine-grained label extraction process. The clinicians systematically mapped the key visual observations (labels) that radiologists describe in the reports and grouped the labels into lexically and semantically meaningful groups based on their visual appearance similarities. Using a top-down approach, the clinicians iteratively searched through the best practice literature including Fleishner Society guidelines[8], consulted several practicing radiologists, and provided a raw list of everyday use terms from their own practices to arrive at a list of core findings in each of the finding type categories. Next, using a bottom-up approach, we mined report collections derived from a variety of data sources, including Indiana dataset[18] (3000 reports), internally labeled collection created from NIH supplied data[3] (16,000 reports), and the MIMIC-4[6] (over 180,000 reports) reports. We extracted frequently occurring n-grams (n varied from 1 to 13) that also had a mapping to meaningful categories relating the UMLS concept categories shown in Table 2. The frequently occurring n-grams were queried against a clinical knowledge base[9]. assembled from 70 reference vocabularies from UMLS. The original knowledge base had over 5.3 million concepts and has been previously reported[9]. This gave rise to about 1500 core terms useful for findings vocabulary generation. To capture and relate the various forms of describing a finding (e.g. infiltrate, infiltration), or alternate ways of saying the same ("Cardiomegaly", "heart is enlarged" and "enlarged cardiac silhouette" ), the clinicians used a glossary development tool[7] and expanded the terminology to cover various abbreviations, misspelling, and semantically equivalent ways of describing the same radiology concepts (synonyms and alternate forms), and ontologically related concepts. Each expansion was reviewed by two radiologists for agreement. Currently, the lexicon consists of over 11000 unique terms covering the space of 78 core findings and 9 modifiers listed in Table 3 which represent the largest set of core finding labels assembled for chest radiographs to date.

*Table 1 The valid modifier associations with the various finding types found in chest radiographs. The | separates the modifiers in the finding label description.*

| Finding Type | Allowed Modifiers for fine-grained labels |
|---|---|
| Anatomical Finding | Anatomicafinding\|present(positive)/absent(negative)\|corefinding\| anatomyaffected\|subanatomy\|location\|laterality\|severity\|size\|hedge\| character\|procedure\|shape\|correlation\|measure\|cause\|symptom |
| Disease | Disease\|present(positive)/absent(negative)\|corefinding\|anatomyaffected\| subanatomy\|location\|laterality\|severity\|size\|hedge\|character\|procedure\| shape\|correlation\|measure\|cause\|symptom |
| Devices | Device\|presentorabsent\|corefinding\|anatomyaffected\|subanatomy\|location\| laterality\|severity\|size\|hedge\|character\|procedure\|shape\|correlation\|measure\| cause\|symptom |
| Tubes and Lines | Tubesandlines\|present (positive)/absent(negative)\|corefinding\|tubesandlines\|subanatomy\| tubesandlineslocation\|laterality\|severity\|size\|hedge\|character\|procedure\| shape\|correlation\|measure\|cause\|symptom |
| Tubes and Lines Finding | Tubesandlinesfinding\|ill-placed (positive)or well-placed (negative)\|corefinding\|anatomyaffected\|subanatomy\|tubesandlineslocation\| laterality\|severity\|size\|hedge\|character\|procedure\|shape\|correlation\|measure\|cause\|symptom |
| Viewpoint | Views\|positive\|finding name |

*Table 2. Illustration of concept categories of UMLS retained as relevant for findings vocabulary generation.*

| | | |
|---|---|---|
| signs and symptoms | acquired abnormality | anatomical |
| congenital abnormality | disease or syndrome | injury or poisoning |
| neoplastic process | anatomical structure | body location or |
| body part, organ, or organ component | body space or junction | body system |
| fully formed anatomical structure | therapeutic or preventive procedure | diagnostic procedure |

## 2.2 Extracting fine-grained labels from radiology reports

The chest X-ray lexicon provided a catalog of core finding labels along with their synonym variants which can now be used to locate them in radiology reports for image labeling purposes. Our approach for fine-grained label generation consists of 4 steps, namely, (a) core finding and modifier detection, (b) phrasal grouping, (c) negation sense detection, (d) pattern completion.

### 2.2.1 Core Findings Detection:

We used a vocabulary-driven concept extraction algorithm to spot all occurrences of core concepts and/or their synonym variants in sentences within reports. For this, we pre-processed the radiology reports to isolate the sections describing the findings and impression. Often, these are indicated by the section headings found in radiology reports. The concept extraction algorithm first builds a vocabulary index in which each synonym variant include the core finding points to the core finding phrase in the lexicon. This ensures that a match to a core phrase can be found through its synonyms. To ensure a match to various word forms of the vocabulary phrases, we pre-process the vocabulary terms by retaining essential prefixes of terms within so that the combined presence of these prefixes points to the actual vocabulary phrase. Table 4, Column 1 lists the prefix strings for the vocabulary phrases in Column 2. The set of prefixes that best discriminate a vocabulary phrase can be determined by the following deterministic algorithm that iteratively shortens each term in a phrase until it fails to be discriminatory in identifying the vocabulary phrase as shown in Figure 2b. The lexicon vocabulary is preprocessed by this smallest prefix building algorithm to record all prefix strings in the vocabulary index. For detecting the vocabulary phrase, all prefix terms from vocabulary phrases are searched within the sentences from relevant sections of reports, and those vocabulary phrases with full matches to their prefixes are retained. This method minimizes the false positives in matching the concepts, particularly for multi-term phrases. Once the candidate vocabulary phrases are identified, a detailed match is initiated within the sentences in which they were found using a dynamic programming algorithm to align the words of candidate vocabulary phrase to the sentence using the prefixes . The resulting alignment guarantees the largest number of words of the vocabulary phrase are matched to the largest possible extent in the sentence while still maintaining the word order and allowing missed and spurious words in between. Given a query vocabulary phrase $S =< s_1 s_2 ... s_K >$ of $K$ words and a candidate sentence $T =< t_1 t_2 ... t_N >$ of $N$ words, we define longest common subfix as $LCF(S,T) =< p_1 p_2 ... p_L >$, where $L$ is the largest subset of words from $S$ that found a partial match in $T$, and $p_i$ is a partial match of a word $s_i \in S$ to a word in $T$. A word $s_i$ in $S$ is said to partially match a word $t_j$ in $T$ if it shares a maximum length common prefix $p_i$ such that $\frac{|p_i|}{\max\{|s_i|,|t_j|\}} \geq \tau$. If we make the threshold = 1.0, this reduces to the case of finding exact matches to words of $S$. We chose to align with prefix in our formulation to correspond to the English grammar rules where many word forms of words share common prefixes. This allows us to model word variants such as regurgitated, regurgitating, and regurgitation as they all share a sufficiently long prefix 'regurgitat'. It can also model spelling errors, particularly those that are made in the later portion of a word which will be deemphasized during alignment. The LCF algorithm is shown in Figure 2a. Here $p_{max}(i,j)$ is the longest prefix of the strings $s_i t_j$ and $\delta$ is a mismatch penalty, which controls the separation between matched words and prevents words that are too far apart in a sentence from being associated with the same vocabulary phrase, thus minimizing the effect of incorrect anaphora resolution in a sentence.

*Table 3. Illustration of core finding labels found sufficient for describing all findings in AP chest radiographs.*

| Category | Finding | Category | Finding | Category | Finding |
|---|---|---|---|---|---|
| anatomical finding | Azygous fissure | anatomical finding | Lobectomy | device | Cardiac pacer and wires |
| anatomical finding | Bone lesion | anatomical finding | Lymph node calcification | device | Msk or spinal hardware |
| anatomical finding | Bullets & foreign bodies | anatomical finding | Mass/nodule (not otherwise specified) | device | Other internal post-surgical material |
| anatomical finding | Calcified nodule | anatomical finding | Mediastinal displacement | device | Sternotomy wires |
| anatomical finding | Consolidation | anatomical finding | Multiple masses/nodules | technical assessment | Apical kyphotic |
| anatomical finding | Contrast in the gi or gu tract | anatomical finding | Normal anatomically | Technical assessment | Apical lordotic |
| anatomical finding | Cyst/bullae | anatomical finding | Not otherwise specified calcification | Technical assessment | Apices not included |
| anatomical finding | Degenerative changes | anatomical finding | Not otherwise specified opacity/pleural/parenchymal opacity | Technical assessment | Costophrenic angle not included |
| anatomical finding | Diffuse osseous irregularity | anatomical finding | Osteotomy changes | Technical assessment | Limited by exposure and/or penetration |
| anatomical finding | Dilated bowel | anatomical finding | Other soft tissue abnormalities | Technical assessment | Limited by motion |
| anatomical finding | Dislocation | anatomical finding | Pleural effusion or thickening | Technical assessment | Low lung volumes |
| anatomical finding | Elevated hemidiaphragm | anatomical finding | Pneumomediastinum | Technical assessment | Lungs obscured by overlying object or structure |
| anatomical finding | Elevated humeral head | anatomical finding | Pneumothorax | Technical assessment | Lungs otherwise not fully included |
| anatomical finding | Enlarged cardiac silhouette | anatomical finding | Post-surgical changes | Technical assessment | Non-diagnostic cxr |
| anatomical finding | Enlarged hilum | anatomical finding | Pulmonary edema/hazy opacity | Technical assessment | Rotated |
| anatomical finding | Fracture | anatomical finding | Scoliosis | Tubes and lines | Central intravascular lines |
| anatomical finding | New fractures | anatomical finding | Shoulder osteoarthritis | Tubes and lines | Drain tubes |
| anatomical finding | Old fractures | anatomical finding | Spinal degenerative changes | Tubes and lines | Enteric tubes |
| anatomical finding | Hernia | anatomical finding | Sub-diaphragmatic air | Tubes and lines | Tubes in the airway |
| anatomical finding | Hydropneumothorax | anatomical finding | Subcutaneous air | Tubes and lines finding | Central intravascular lines: incorrectly positioned |
| anatomical finding | Hyperaeration | anatomical finding | Superior mediastinal mass enlargement | Tubes and lines finding | Coiled/kinked/fractured |
| anatomical finding | Increased reticular markings/ild pattern | anatomical finding | Tortuous aorta | Tubes and lines finding | Enteric tubes: incorrectly positioned |
| anatomical finding | Linear/patchy atelectasis | anatomical finding | Vascular calcification | Tubes and lines finding | Tubes in the airway: incorrectly positioned |
| anatomical finding | Lobar/segmental collapse | anatomical finding | Vascular redistribution | | |
| **Modifiers** | | | | | |
| Anatomy affected, subanatomy, location, laterality, severity, size, hedge, character, procedure, shape, correlation, measure, cause, symptom | | | | | |

*Table 4. Illustration of prefix extraction for terms within a vocabulary phrase to increase specificity of matching.*

| Prefix strings | Vocabulary phrase | Matching sentence in a textual report |
|---|---|---|
| Aort:sclero, Aort:sten | Aortic sclerosis, Aortic stenosis | Marked aortic sclerosis present with evidence of stenosis. |
| Frac:clav | Fracture of clavicle | There is a transverse fracture of the mild left clavicle with mild superior angulation of the fracture fragment. |
| Perfor:esop | Perforation of esophagus | A contrast esophagram shows esophageal perforation of the anterior left esophagus at C4-5 with extraluminal contrast seen. |
| Edem:lower:extrem | Edema of lower extremity | EXTREMITIES: Lower extremity trace pitting edema and bilateral lower extremity toe ulceration and onychomycosis, right plantar eschar. |
| Atri:dila | Atrial dilatation | Left Atrium: Left atrial size is mildly dilated. |
| Mass:left:brea | Mass in left breast | new lft breast palp mass found. |

Using this algorithm, a vocabulary phrase $S$ is said to be detected in a sentence $T$ if $\frac{|LCF(S,T)|}{|S|} \geq \Gamma$ for some threshold $\Gamma$. The choice of $\tau$ and $\Gamma$ affect precision and recall in matching and can be suitably chosen to meet specified criteria for precision and recall based on an ROC curve analysis as is popular in information retrieval literature. Note that the normalization in the above equation is on the length of the vocabulary phrase and not the sentence allowing matches to be found in long sentences.

```
LCF(S,T);
C[i, 0] = 0, C[0, j] = 0,  0 ≤ i ≤ K, 0 ≤ j ≤ N
for (1≤ i ≤ K)
  for (1≤ j ≤ N)
  {
    ρ_ij = |p_max(i,j)| / max{|s_i|,|t_j|} ;
    If C[i-1,j-1] + ρ_ij > C[i-1,j] && C[i-1, j-1] + ρ_ij > C[i,j-1]
       C[i,j] = C[i-1,j-1] + ρ_ij;
    Else
    {
      If C[i-1,j] + ρ_ij > C[i,j-1]
         C[i,j] = C[i-1,j] - δ;
      Else
         C[i,j] = C[i,j-1] - δ;
    }
  }
```

```
findSmallestForm (word)
{
    found = false;
    i = word.length();
    prefix = word;
    while (!found && i >= 3)
    {
       prefix = word.substring(0,i);
       if ((prefix not in wordMap) || (prefix not shared in wordMap))
          i--; // continue shrinking
       else
          found = true;
          prefix = word.substring(0, i + 1)
    }
    return prefix;
}
```

*Figure 2. Illustration of (a) longest common subfix algorithm, (b) smallest distinguishable prefix formation algorithm per word in phrase.*

Table 4, Column 2 shows the vocabulary phrases that were recognized from the sentences shown in Column 3 of that table. As can be seen, the algorithm was able to spot the occurrence of both 'aortic sclerosis' and 'aortic stenosis' in the sentence, even though the words 'aortic' and 'stenosis' are separated by several words in between. Similarly, the vocabulary phrase 'left atrial dilatation' was matched to 'Left Atrium: Left atrial size is mildly dilated' even without deep understanding of the linguistic origins of the underlying words.

### 2.2.2 Association of modifiers with relevant core findings:

The above vocabulary-driven phrasal detection algorithm can be applied to the vocabulary of both core findings and modifiers to appropriately tag phrases within sentences. To generate the fine grained finding label, we need to associate the modifiers with the relevant findings. Doing this without full natural language understanding can be difficult. For example, in the sentence "The lungs are normally inflated without evidence of focal airspace disease, pleural effusion or pneumothorax," is the modifier "focal" associated with airspace disease only or also with pleural effusion and pneumothorax? The approach we used was to employ a natural language parser called the ESG parser[10] which performs word tokenization, sentence segmentation, morpho-lexical analysis, and syntactic analysis to produce a dependency parse tree called the SG parse tree. In SG parse tree, each tree node N is centered on a headword, which is surrounded by its left and right modifiers, which are, in turn, tree nodes. Each modifier M of N fills a slot in N. The slot shows the grammatical role of M in N and is indicated by a tuple T=(t1,t2,…tk) which means that t1 is a term grammatically related to modifiers t2,…tk. Here, an unknown modifier is indicated by the symbol 'u'. A sample SG parse tree for the sentence "The lungs are normally inflated without evidence of focal airspace disease pleural effusion or pneumothorax" is shown in Figure 3a. The association tuples are also shown in this figure such as for the word "without" through (6,5,7) indicating the word "without" is relating term "inflate" to "evidence". Associations that logically go together such as adjectives describing nouns, are already indicated by the ESG parser through numeric codes exceeding 100, such as for the term "pleural effusion" which has the slot structure (211) and is also seen by the pairing (12,13).

Given such a dependency parse tree G and the tuples $T_G = <T_1, T_2, ... T_N>$ corresponding to the N tree nodes, where $T_i = (t_{i1}, .. t_{k_i})$ are the tuple per node, we define a phrasal group as $P_l = (e_1, e_2, .. e_M)$
where $e_j = t_{jk} \in T_j$ is the kth element of some tuple $T_j$ and $\forall_{j=1}^{M} T_j \cap T_{j+1} \neq \emptyset$. In other words, a phrasal group is a connected component formed from the transitive closure of the tuples such that they have at least one element in common. Figure 3b illustrates the phrasal grouping process and the groups produced for the sentence shown at the top of the figure. Since the core findings and modifiers are detected from a prior stage, they can now be mapped back into the phrasal group. Phrasal groups that contains one or more core findings are called core phrasal groups while the rest of the groups are called the helper groups. If a core finding is detected across two or more adjacent core groups, then they are also merged to form a single core group as shown in Figure 3b (the blue-colored group). All modifiers present in a core phrasal group are associated with the corresponding core findings. Finally all modifiers present in helper groups are associated with the core findings of their adjacent groups. Figure 3b also lists the various phrasal groups and the two core finding associations found in the sentence (shown by blue arcs).

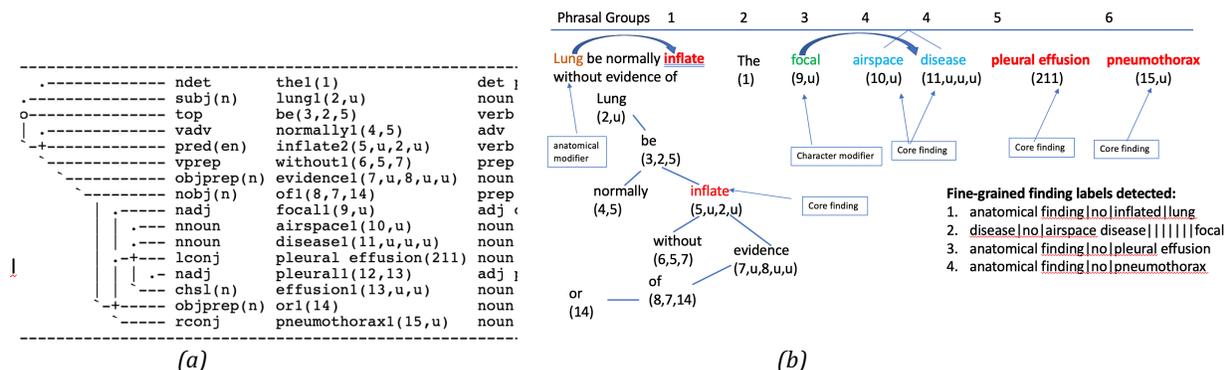

*Figure 3: Illustration of phrasal group extraction. Dependency parse tree of the sentence above generated by the ESG parser. (b) The phrasal grouping process using connected component analysis. The core findings from the lexicon are highlighted in red that occur within phrasal groups. Those that cross phrasal groups are indicated in blue. The character modifiers from lexicon detected are shown in green.*

### 2.2.3 Negated instance detection of core findings:

To determine if a core finding is a positive or negative finding (e.g. "no pneumothorax"), we use a two-step approach that is a combination of language structuring and vocabulary-based negation detection extending a previous work[9]. The language structuring approach to negation detection starts from a dependency parse tree of a sentence. A set of known typed dependency patterns developed by the Stanford NLP parser team[9] were used to search for negations and the scope of words spanned by a negation keyword. The negation pattern detection algorithm iteratively identifies words within the scope of negation based on dependency parsing. Let $S$ be the set of negated words. The algorithm starts by adding a collection of manually curated negation cues (e.g. 'no') into $S$, and then iteratively expand $S$ through traversing the dependency parse tree of a sentence, until $S$ becomes stable. Figure 4 illustrates negation detection. Based on the language analysis of the sentence and the negation pattern matches, the negation scope is listed as : 'evidence', 'suggesting', 'has', and 'cancer', and the target vocabulary phrase is identified as 'cancer'.

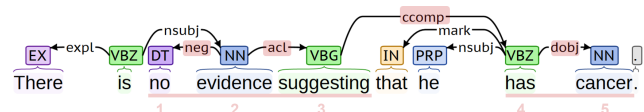

*Figure 4: Illustration of negation detection*

The above negation detection algorithm is dependent on the correctness of the dependency parse tree. To ensure that the negation modifiers are being associated with the relevant core phrase, we also formed a vocabulary of "negation prior" and "negation post" terms such that their occurrence prior or post the core finding is a further indication of negation or avoiding spurious negation detection. The language-based negation detection algorithm was previously compared against comparable approaches and shown to be superior to NEGEX on large collections[9]. By adding the pre and post negation terms from the chest X-ray lexicon, we were able to improve the performance of the negation detector further by 10%. As we will show in the results section, the error in labeling due to negation errors is now less than 2%.

**2.2.4 Fine-grained finding pattern formation:** To form the fine-grained finding label using the label pattern syntax, we begin with the core finding and the associated modifiers discovered from the phrase group. For each core finding, its finding type is retrieved from the chest X-ray lexicon. Further, due to the a priori knowledge captured in the lexicon for the associated anatomical locations of findings we can augment the fine-grained label pattern with the anatomical location even when these are not specified in the sentence itself. In addition, the name of the core finding may be ontologically rolled up to the core findings from the chest X-ray lexicon.

Figure 3b lists the fine grained labels extracted from the sentence shown in that figure. As can be seen, both positive and negative instances of findings have been extracted by this process. Table 5 lists examples of types of fine-grained labels extracted from sentences from redacted reports. As can be seen, important details of the finding are adequately captured in the generated fine-grained label. By mining the Findings and Impression of over 220,000 radiology reports from our collection using the above method, we recorded all possible fine grained labels that could be extracted. By retaining only those labels with at least 100 image support needed for deep learning model construction, a total of 457 fine-grained labels were selected. Of these, 78 were the original core labels given by clinicians shown in Table 3, and the remaining were finer description labels with modifiers extracted automatically. The list of all 457 labels was too long to list in the paper but can be furnished upon request.

*Table 5. Examples of fine-grained labels extracted from sample sentences.*

| Sentence | Fine-grained label | Semantics | |
|---|---|---|---|
| there is left base streaky opacity due to xxxx scarring or discoid atelectasis. | anatomicalfinding\|yes\|discoid atelectasis\|\|\|\|\|\|\|\|\|discoid | Findingtype=anatomical finding<br>Core finding=discoid atelectasis<br>Positive finding? yes<br>Shape modifier=discoid | |
| there is left base streaky opacity due to xxxx scarring or discoid atelectasis. | anatomicalfinding\|yes\|streaky opacity\|base\|\|left;;base\|left\|\|\|or\|\|\|streaky | Findingtype=anatomical finding<br>Core finding=streaky opacity<br>Anatomy=base<br>Location=left;;base | Laterality=left<br>Character=streaky<br>Positive finding? yes |
| there is left base streaky opacity due to xxxx scarring or discoid atelectasis. | anatomicalfinding\|yes\|scarring | Findingtype=anatomical finding<br>Core finding=scarring<br>Positive finding? yes | |
| the right upper extremity picc tip is in the upper svc. | tubesandlinesfinding\|yes\|upper svc\|\|\|upper svc | Findingtype=position of tubes and lines<br>Isapositive findings? yes<br>Corefinding=upper svc (happens to be an anatomical structure, this is a case where the core finding is the anatomy itself)<br>Location=upper svc | |
| the right upper extremity picc tip is in the upper svc. | tubesandlines\|yes\|picc\|extremity upper;;right upper extremity\|\|right\|right | Findingtype=tubes and lines<br>Positivefinding? yes<br>Core finding=picc (line) | Subanatomy=right upper extremity<br>tubesandlineslocation=right<br>Laterality=right |

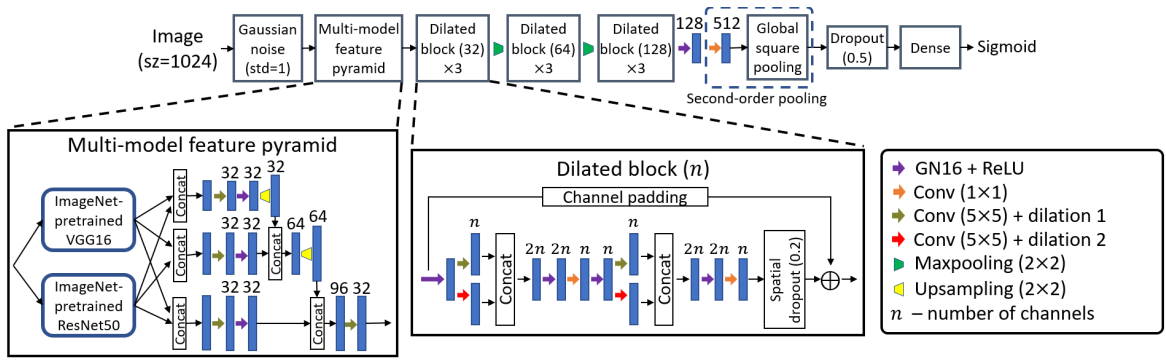

*Figure 5: The custom deep learning network built for fine-grained label classification.*

## 2.3 Building deep learning models with fine-grained labels

We now describe a deep learning model developed to distinguish between the fine-grained labels. The learning of such labels from chest radiographic images is a fine-grained classification problem for which single networks used for computer vision problems may not yield the best performance as large training sets are still difficult to obtain. Concatenating different ImageNet-pretrained features from different networks can improve classification as was shown for microscopic images[11]. Following this idea, we combine the ImageNet-pretrained features from different models through the Feature Pyramid Network[12] using features across multiple scales. Specifically, the VGGNet (16 layers) [13] and ResNet (50 layers) [14] are used as the feature extractors and their lower-level features are retained. In particular, from the VGGNet, the feature maps with 128, 256, and 512 feature channels are used, which are concatenated with the feature maps from the ResNet of the same spatial sizes which have 256, 512, and 1024 feature channels. Dilated blocks are used to learn the high-level features from the extracted ImageNet features. Each dilated block is composed of dilated convolutions for multi-scale features[15], and uses a skip connection of identity mapping to improve convergence[16] and spatial dropout to reduce overfitting. Group normalization (16 groups) is also used with ReLU. Dilated blocks with different feature channels are cascaded with max pooling to learn more abstract features. Second-order pooling is used, which is proven to be effective for fine-grained classification[17] and maps the features to a higher-dimensional space where they are more separable. The resulting network design used for multi-label finding pattern classification is shown in Figure 5.

## 3. Results

Since our goal was to produce a deep learning network trained for fine-grained description of a wide spectrum of findings in AP chest X-rays, we evaluate our approach along 3 directions, namely, (a) extent of coverage of findings, (b) accuracy of the fine-grained label extraction algorithm, and (c) finally, the performance of the deep learning network.

### 3.1 Extent of coverage:
To demonstrate wide coverage of all fine-grained findings found in AP chest radiographs, we collected datasets from 3 separate sources, namely, the MIMIC-4[6], a dataset of over 220,000 reports with associated frontal images, 2964 unique Indiana reports[18] with associated images, and a set of 10,000 NIH[3] released images re-read by our team of radiologists to produce a total of 232,964 image-report pairs for our analysis. A total of 203,938 unique sentences were processed resulting in 102,135 fine-grained labels. We recorded 91% of the sentences to contain one or more findings, and the remaining were sentences that were either non-indicative sentences, such as for example, "chest comparison." (6% cases), or those missed due to inadequate coverage of synonyms in the lexicon for core findings (4% cases). This showed that in the tested collection of reports, our coverage of findings was adequate.

Next, to test the coverage provided by the fine grained labels we mapped all 102,135 labels derived to the nearest label in the selected 457 label set through a two-step process, namely, mapping to a core finding in the lexicon via synonym lookup, or to core findings via ontological concept expansion from the lexicon. For example, all emphysema-related labels could be mapped to "cysts/bullae" concept category in our lexicon. Since the fine-grained labels were seeded by clinically selected core findings, nearly 83% of all labels extracted could be mapped into their nearest counterpart in the 457 fine-grained label set. The remaining 13% of the case had lower than 1% incidence statistics and data so that deep learning models could not be built for those cases anyway. Thus for all practical purposes, the set of core findings and the 457 fine-grained labels were found sufficient to cover a wide variety of findings occurring in radiology reports.

*Table 6. Results of testing the accuracy of fine-grained label extraction.*

| Reports Analyzed | Relevant sentences | Fine-grained labels | Missed findings | Overcalled findings | Negation sense errors | Incorrect association | Missed association |
|---|---|---|---|---|---|---|---|
| 2964 | 3046 | 5245 | 0 | 4 | 168 | 49 | 11 |

### 3.1 Accuracy of fine-grained label extraction:
Since we used a vocabulary-driven approach to extract the core findings in the label extraction, we evaluated the correctness of the finding detection in terms of misses and overcalls on findings. We also noted the correctness of the positive or negative instantiation of the core finding. Finally, we evaluated the correctness of association of the modifiers to the core findings as generated through the phrasal grouping algorithm. The Indiana dataset[18] was used for this purpose. Specifically we collected sentences from the Findings and Impression sections of 2964 reports corresponding to all unique frontal view AP chest radiographs from this collection. A total of 5245 sentences were examined by clinical experts to catalog the number of misses, overcalls of core findings and their associations, one sentence at a time. The results are shown in in Table 6. As can be seen from this table, the label extraction process is highly accurate in the detection of core findings and the majority of the errors are in capturing the negation sense which is around 3%. Further, we can see that the association of modifiers to core findings given by the phrasal grouping algorithm is also highly accurate with over 99% precision and recall (49 of the 5245 labels).

## 3.2. Deep learning model evaluation

For purposes of training the deep learning models, we used the image data from MIMIC-4[6] and NIH[3] datasets and a training-validation-test split of 80%-10%-20% respectively as shown in Table 7. Using the Nadam optimizer for training with a learning rate of $2\times10^{-6}$, a batch size of 48, and 20 epochs, the average area under the curve across all 457 labels was 0.73. By focusing only on the correctness of the core findings only, a larger dataset was possible for training, validation and testing as shown in the upper row, and the resulting unweighted AUCs were higher at 0.81 as shown in Table 7.

*Table 7. Results of testing the accuracy of fine-grained label classification by the deep learned model.*

| Dataset | Train | Validate | Test set | Average AUC | Weighted Average AUC |
|---|---|---|---|---|---|
| MIMIC-4 + NIH | 249,286 | 35,822 | 70,932 | 0.81 | 0.84 |
| MIMIC-4 + NIH | 75613 | 10,615 | 20,941 | 0.73 | 0.73 |

The performance varied per fine-grained label based on the number of images available for training as well as the complexity implied by the label. Figure 6 shows the results of classification accuracy for a spectrum of fine-grained findings and their associated ROC curves. Figure 7 shows examples of images accurately classified by the deep learning model reflecting the nuances of laterality and location details captured by image recognition. Even uncertain findings called hedges can be detected by the network as shown by the performance on the label "slight pulmonary edema".

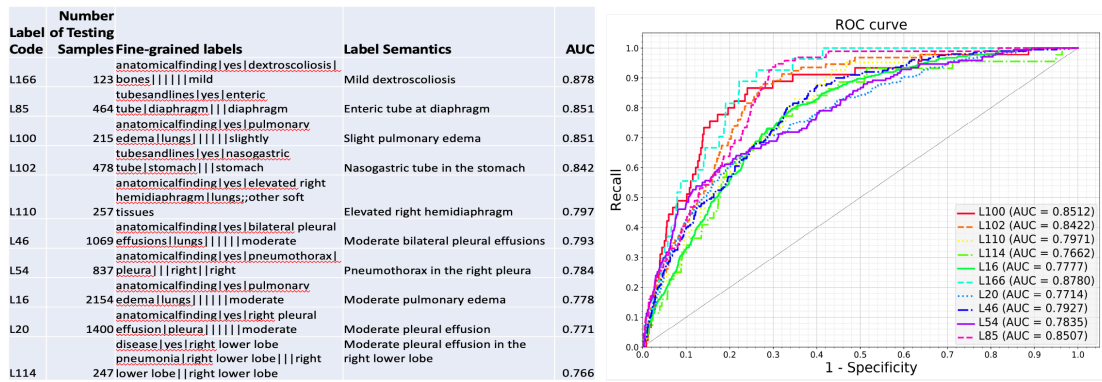

*Figure 6: Illustration of performance of the deep learning network for a variety of fine-grained labels.*

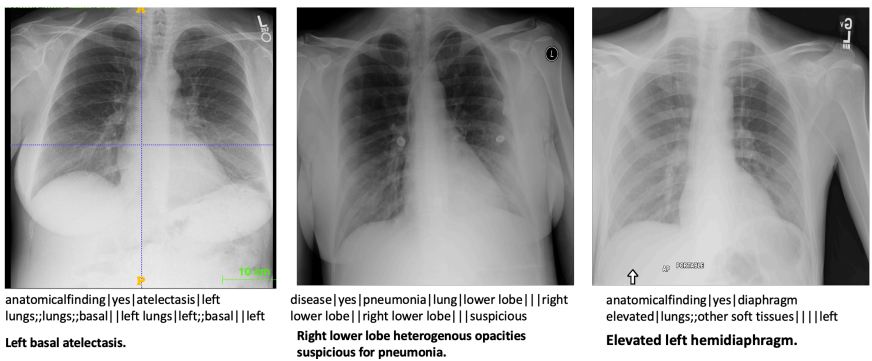

*Figure 7: Illustration of the variety of findings that are accurately classified by the deep learning model.*

## 4. Discussion & Conclusions

Deep learning has become the method of choice for pattern recognition in medical image analysis. The quality of learning, however, is a function of the granularity of labels that can be attached to images. In this paper, we present a new method for extracting fine-grained labels from images and use it to train a deep learning model to recognize fine-grained descriptions of findings. Although the overall results show that it is possible to automatically label and learn fine-grained labels using pattern recognition approaches of deep learning models, the performance obtained is still dependent on the amount of training data per label which can reduce as we increase the specificity of the labels. Further, the accuracy in labeling could be further improved through increased negation vocabulary addition. Future work will also incorporate regional information directly into the learning model to compensate for the reduced datasets for the labels.